%% file: main.tex
\pdfoutput=1
\documentclass[]{spie}

\usepackage{amsmath,amsfonts,amssymb}
\usepackage{graphicx}
\usepackage{subfig}
\usepackage[export]{adjustbox}
\usepackage{booktabs}
\usepackage{todonotes}

\usepackage[colorlinks=true, allcolors=blue]{hyperref}
\usepackage{titlesec}
\usepackage{floatrow}
\newfloatcommand{capbtabbox}{table}[][\FBwidth]

\definecolor{mo-so-elastix}{HTML}{ff6f6b}
\definecolor{mo-elastix}{HTML}{5388ff}
\definecolor{morea}{HTML}{4ed216}
\newcommand{\mosoelastix}{\textcolor{mo-so-elastix}{red}}
\newcommand{\moelastix}{\textcolor{mo-elastix}{blue}}
\newcommand{\morea}{\textcolor{morea}{green}}

\newcommand{\plotwidth}{0.97}

\usepackage{varwidth}
\DeclareCaptionFormat{myformat}{%
  \begin{varwidth}{\linewidth}%
    \centering
    #1#2#3%
  \end{varwidth}%
}

\title{A Tournament of Transformation Models: \\ B-Spline-based vs.\ Mesh-based \\ Multi-Objective Deformable Image Registration}

\author[a]{Georgios Andreadis}
\author[b]{Joas I. Mulder}
\author[c]{Anton Bouter}
\author[c,b]{Peter A. N. Bosman}
\author[a]{\\Tanja Alderliesten}
\affil[a]{Dept. of Radiation Oncology, Leiden University Medical Center (LUMC), P.O. Box 9600, 2300 RC Leiden, The Netherlands}
\affil[b]{Faculty of Electrical Engineering, Mathematics and Computer Science, Delft University of Technology, Van Mourik Broekmanweg 6, 2628 XE Delft, The Netherlands}
\affil[c]{Evolutionary Intelligence Group, Centrum Wiskunde \& Informatica (CWI), P.O. Box 94079, 1090 GB Amsterdam, The Netherlands}

\authorinfo{Address all correspondence to: Georgios Andreadis (G.Andreadis@lumc.nl) and/or Tanja Alderliesten (T.Alderliesten@lumc.nl).}

\begin{document} 
\maketitle

\input{sections/0-abstract}


\input{sections/1-introduction}
\input{sections/2-materials-and-methods}

\input{sections/3-results}
\input{sections/4-discussion-and-conclusions}

\acknowledgments 
 
The authors thank W. Visser-Groot and S.M. de Boer MD, PhD (Dept. of Radiation Oncology, Leiden University Medical Center, Leiden, The Netherlands) for their contributions to this study.
This research is part of the research programme Open Technology Programme with project number 15586, which is financed by the Dutch Research Council (NWO), Elekta, and Xomnia. 
Further, this work is co-funded by the public-private partnership allowance for top consortia for knowledge and innovation (TKIs) from the Ministry of Economic Affairs.

\bibliography{references} 
\bibliographystyle{spiebib} 

\end{document}

%% file: sections/0-abstract.tex
\begin{abstract}
\noindent
The transformation model is an essential component of any deformable image registration approach.
It provides a representation of physical deformations between images, thereby defining the range and realism of registrations that can be found.
Two types of transformation models have emerged as popular choices: B-spline models and mesh models.
Although both models have been investigated in detail, a direct comparison has not yet been made, since the models are optimized using very different optimization methods in practice.
B-spline models are predominantly optimized using gradient-descent methods, while mesh models are typically optimized using finite-element method solvers or evolutionary algorithms.
Multi-objective optimization methods, which aim to find a diverse set of high-quality trade-off registrations, are increasingly acknowledged to be important in deformable image registration.
Since these methods search for a diverse set of registrations, they can provide a more complete picture of the capabilities of different transformation models, making them suitable for a comparison of models.
In this work, we conduct the first direct comparison between B-spline and mesh transformation models, by optimizing both models with the same state-of-the-art multi-objective optimization method, the Multi-Objective Real-Valued Gene-pool Optimal Mixing Evolutionary Algorithm (MO-RV-GOMEA).
The combination with B-spline transformation models, moreover, is novel.
We experimentally compare both models on two different registration problems that are both based on pelvic CT scans of cervical cancer patients, featuring large deformations.
Our results, on three cervical cancer patients, indicate that the choice of transformation model can have a profound impact on the diversity and quality of achieved registration outcomes.
\end{abstract}

%% file: sections/1-introduction.tex
\section{INTRODUCTION}

Many image-guided clinical treatments could benefit from the transfer of information between multiple medical imaging scans of the same patient.
This transfer of information can be hampered by, e.g., local deformations of patient tissue or content mismatch between the images, calling for a deformable image registration (DIR) approach that can find a spatial correspondence between two images.
An important component of any such DIR approach is its transformation model, which holds assumptions about how physical deformations occur~\cite{Rueckert2011}.
The choice of transformation model therefore influences the range and quality of registrations that can be found by using that DIR approach.
Two commonly used free-form transformation models in DIR are B-spline models and mesh models.
B-spline models capture deformations through a composition of B-splines.
Their generality, computational tractability, and smoothness motivate usage in a variety of DIR applications~\cite{Metz2011,Rueckert1999}. 
Mesh models typically express deformations by manipulating tetrahedral meshes, using techniques from finite element modeling (FEM).
Their level of realism and versatility in modeling different tissue types motivate applications to regions with complex deformations~\cite{Brock2005,Rigaud2019}.
Since B-spline and mesh transformation models differ in many aspects, important insights could be gained by comparing the models directly.

The two transformation models are in practice, however, optimized with different optimization methods, which complicates any direct comparison of transformation models between registration approaches.
While B-spline models tend to be optimized using gradient-descent methods\cite{Klein2010,Sharp2010}, mesh models are often optimized using FEM solvers such as preconditioned conjugate gradient algorithms, for forward simulation together with some other algorithm for boundary condition optimization\cite{Brock2005}, or evolutionary algorithms (EAs), for direct optimization~\cite{Andreadis2023}.
Gradient-descent methods are known to converge quickly, but also to be prone to converge to local optima, which can prevent these methods from fully leveraging the potential of the transformation model.
EAs can be slower to converge, but are often less likely to get stuck in local optima, especially when there are many parameters to optimize.
Moreover, they can be naturally and efficiently used for multi-objective optimization~\cite{Deb2001}, meaning that they can return in a single run of the EA a so-called approximation set, consisting of multiple alternative registration solutions with different trade-offs between the underlying optimization objectives.
The diversity of registrations in this approximation set allows multi-objective optimization methods to provide a more complete picture of the capabilities of a transformation model.
Although both the configurations of individual DIR transformation models~\cite{Sun2013,Klein2010,Rigaud2019} and the differences between complete DIR approaches~\cite{Murphy2011,Loi2018} have been studied extensively in previous work, a direct comparison, where transformation models are optimized using the same optimization method, has not yet been made.

This work presents the first direct comparison of DIR transformation models, using the same multi-objective optimization method for two different transformation models.
The Multi-Objective Real-Valued Gene-pool Optimal Mixing Evolutionary Algorithm (MO-RV-GOMEA) has proven effective in optimizing a mesh transformation model within the MOREA registration approach~\cite{Andreadis2023}.
Since using mesh models presents a complex, discontinuous optimization landscape, and the DIR problem has been shown to be inherently multi-objective~\cite{Pirpinia2017}, the MO-RV-GOMEA optimization method is a suitable candidate to optimize both transformation models.
To this end, we introduce a novel, multi-objective B-spline DIR approach optimized with MO-RV-GOMEA.
This enables a true empirical comparison of both B-spline and mesh models, since both are now manipulated by the same optimization method.
To measure the impact of the choice of optimization method, we include a gradient-descent baseline which uses repeated single-objective gradient-descent optimization to multi-objectively optimize objective weights~\cite{Pirpinia2017}.
We compare all three approaches on six clinical registration problems derived from scans of three cervical cancer patients, analyzing the quality and diversity of registrations found by each approach.

%% file: sections/2-materials-and-methods.tex
\section{MATERIALS AND METHODS}

\input{figures/illustrations/decomposition}

\subsection{Transformation models}

To register a source image $I_s$ with a target image $I_t$, a realistic spatial correspondence needs to be found between the two images.
Such a correspondence is defined by an inverse transformation $T'$ which matches each voxel $p_t$ of the target image space to a voxel position $T'(p_t)$ in source image space, such that a resampled source image $T'(I_s)$ approximates $I_t$~\cite{Brown1992}.
This is the direction of transformation typically produced by DIR approaches.
Any transformation uses an underlying transformation model that provides the degrees of freedom in which the transformation can be expressed.
We consider two popular transformation models in this work.

\paragraph{B-spline transformation model}

The B-spline transformation model consists of a fixed set of control points on the target image, and corresponding, movable points on the source image.
These control points form a grid of image patches and parameterize local deformations using generalized Bézier curves.
As shown in Figure~\ref{fig:decomposition:spline-patch}, each patch is controlled by a set of surrounding control points.
By definition, transformations expressed in this model are smooth, but not guaranteed to be free of folds.

\paragraph{Dual-dynamic mesh transformation model}

The dual-dynamic mesh transformation model considered in this work uses triangular meshes (in 2D) or tetrahedral meshes (in 3D) on both images to express transformations~\cite{Alderliesten2013,Andreadis2022}.
Unlike in the B-spline model, the points of both the source and the target mesh are movable, allowing the model to naturally capture (dis)appearing structures and large deformations.
On 3D images, each tetrahedron in the target mesh (see Figure~\ref{fig:decomposition:mesh-patch}) has a corresponding tetrahedron in the source mesh, defining the spatial correspondence of the two image regions that the tetrahedra cover.
The transformations expressed in this model are defined in a piecewise linear fashion (i.e., per tetrahedron), but as the meshes are checked for folds, the resulting transformation is guaranteed to be fold-free.

\subsection{Optimization with MO-RV-GOMEA}
We optimize both transformation models using the same state-of-the-art, multi-objective optimization method: MO-RV-GOMEA~\cite{Bouter2017b}.
A key advantage of MO-RV-GOMEA over standard EAs is that it can leverage partial evaluations, i.e., when local changes in a solution can be evaluated quickly, which is the case here.
For the B-spline transformation model, we introduce a novel registration approach which integrates MO-RV-GOMEA into the Elastix image registration toolbox~\cite{Klein2010}.
This is the first approach that decomposes the B-spline transformation model into local regions, enabling simultaneous, partial optimization of non-neighboring image regions.
As Figure~\ref{fig:decomposition:spline-dependencies} illustrates, when one control point is manipulated, only the surrounding patches it affects need to be recomputed.
Coefficients (one for each dimension) are grouped per control point and modified together in each partial change.

For the mesh transformation model, we use the existing MOREA registration approach~\cite{Andreadis2023}, which already uses MO-RV-GOMEA to optimize a custom initialized dual-dynamic mesh with hardware-acceleration on the Graphics Processing Unit (GPU).
In this approach, all coordinates of mesh points connected by an edge are modified together.
A subset of edges to modify is automatically selected such that each mesh point is included in at least one selected edge.

We also include a comparison baseline, to measure the impact of the choice of optimization method on B-spline-based registration.
This baseline uses MO-RV-GOMEA to optimize the objective weights of the original single-objective, gradient-descent-based Elastix B-spline registration approach.
The weights of both optimization objectives are manipulated together.
A specific combination of optimization objective values is then evaluated by executing a full run of the Elastix toolbox with given objective weights.
This is similar to earlier work on black-box multi-objective B-spline registration~\cite{Pirpinia2017}, although with a different, more effective optimization method.

\input{figures/illustrations/optimization-methods}

\subsection{Optimization objectives}

Registrations are evaluated using two objectives during optimization, concerning their image similarity and deformation magnitude.

\paragraph{Image similarity objective}

This objective assesses how well the transformed source image $T'(I_s)$ matches the target image $I_t$, visually.
We use the \textit{sum of squared differences} metric to capture differences between the image intensities of sampled (interpolated) target image voxel positions $p_t \in P_t$ and their corresponding (interpolated) source image voxel positions $T'(p_t)$.
Both models are configured to randomly sample the source and target image space at $|P_t|$ voxel positions, chosen to be equal to the number of discrete voxels present in the target image.
The objective is defined as follows:
\[
f_{\text{\emph{similarity}}} = \frac{1}{|P_t|} \sum_{p_t \in P_t} (I_t(p_t) - I_s(T'(p_t)))^2
\]

\paragraph{Deformation magnitude objective}

This objective describes the amount of energy needed to perform the transformation.
For each model, we use a different, generally accepted formulation of this energy:
The B-spline model estimates \textit{thin-plate bending energy} from control point coefficients, while the mesh model employs \textit{Hooke's law} to assign energy to differences in edge lengths between the two meshes.

The objective formulation of the B-spline model samples the inverse transformation $T'$ at $|P_t|$ points and computes the gradient of the transformation at each point~\cite{Klein2010}.
Given the second derivatives of the transformation w.r.t. a variable voxel position $x$, the objective is defined as follows:
\[
f_{\text{\emph{def. magnitude: B-spline}}} = \frac{1}{|P_t|} \sum_{p_t \in P_t}
\left\| \frac{\partial^2 T'}{\partial x \partial x^T}(p_t)
\right\|^2_F
\]

The objective formulation of the mesh model considers all corresponding source mesh edges $e_s$ and target mesh edges $e_t$ of each tetrahedron $\delta \in \Delta$.
For each tetrahedron, it also includes 4 spoke edges that better capture flattening motion, giving a total of 10 edges~\cite{Andreadis2022}.
Using Hooke's law~\cite{Arfken1985}, the objective is defined as follows:
\[
f_{\text{\emph{def. magnitude: mesh}}} = \frac{1}{10 |\Delta|} \sum_{\delta \in \Delta} \left[ \sum_{(e_s,e_t) \in E_\delta} (\lVert e_s \rVert - \lVert e_t \rVert)^2 \right] 
\]

\subsection{Experiments}

For an empirical comparison of the two transformation models, we consider three aspects in our experimental setup: 1) the configuration of the models and associated optimization methods; 2) the registration problems; and 3) the method of comparison.

\paragraph{Configuration}

The number of parameters of a model determines at what level of detail deformations can be modeled.
In our comparison, this number is set to be approximately equal for both transformation models.
For the B-spline model, a control point grid with $7 \times 7 \times 7$ points is chosen (corresponding to $7^3 \times 3 = 1029$ variables), and for MOREA, a custom surface-based mesh is generated with 170 points for each image ($170 \times 3 \times 2 = 1020$ variables).
We use single-resolution registration for both transformation models, and perform registration without using masks.
Each model is optimized with the same parameter settings of MO-RV-GOMEA:
Across objective space, 10 clusters of solutions are optimized simultaneously, and a target capacity of 1000 is used for the elitist archive, used for storing high-quality registration solutions.
The baseline DIR approach, which executes single-objective runs of B-spline registration and manipulates their objective weights in a black-box fashion, is optimized with MO-RV-GOMEA with the same above-mentioned configuration.
To make this approach computationally tractable, a random sampling strategy with 5000 sample points and a maximum of 2000 iterations is set for the internal, single-objective Elastix optimization.

All approaches are checked for convergence before reporting results.
Population sizes are chosen to facilitate this convergence, while maintaining computational tractability: the baseline approach is executed with a population size of 100, the multi-objective B-spline approach with a population size of 200, and the multi-objective mesh approach with a population size of 700.
In practice, using 200 generations results in convergence in all three tested DIR approaches.
All experiments are repeated 10 times for reproducibility.
For visualization purposes, one repetition is automatically chosen from these repetitions by computing the hypervolumes~\cite{Zitzler2008} (a quality indicator of approximation sets) of all repetitions and selecting the repetition with the median hypervolume.

\paragraph{Problems}

We derive our registration problems from CT scans of the pelvic area of three cervical cancer patients, acquired for radiation treatment planning purposes, using a Brilliance Big Bore scanner (Philips Healthcare).
The source image pictures a full bladder, and the target image, taken shortly thereafter, pictures an empty bladder.
Both images had in-slice resolutions ranging from 0.86mm to 1.08mm, and a slice thickness of 3mm.
These images are then resampled to a common resolution of $(3,3,3)$\textit{mm} and rigidly registered to align the bony anatomies, using bone contours delineated by a radiation therapy technologist~(RTT), for consistency.
The images are then cropped to an axis-aligned bounding box around the bladder with a 30\textit{mm} margin, using the maximal bounds from the source and target image.

\input{figures/results/p1/problem}

Registration problems in this comparison should be solvable without the use of additional contour guidance, since the B-spline-based approach does not support this as an objective, and the comparison should be fair to both models.
Earlier work shows that B-spline models struggle to register pelvic image regions with low contrast without pre-segmenting the images provided to the model~\cite{Andreadis2023}.
To accommodate for this, we derive two registration problems for each patient that implicitly provide segmentations but still preserve image intensity values within organs.
Slices of these two registration problems are shown in Figure~\ref{fig:problem}.
The first problem we derive, consists of only the bladder, isolated from the rest of the anatomy captured in the pelvic scan.
This features a large deformation, since the bladder varies strongly in filling between scans.
An implicit segmentation contour is provided by the sharp contrast between the (gray) intensity values of the bladder and the (black) isolated background.
The second problem consists of a more complex registration problem: the bones, sigmoid, and rectum are added to the image, similarly isolated from the rest of the body and organs.
The bowel is not added to this problem, since the closeness of bladder and bowel proved to be difficult to register without supplying contour guidance.

\paragraph{Comparison}

All registrations found are re-evaluated in a common pipeline, in the interest of fairness:
Each registration is first rasterized in the form of a deformation vector field (DVF), and then evaluated on both optimization objectives.
Since the two formulations of the deformation magnitude objective differ, we re-evaluate all registrations afterward using a voxel-based metric, by calculating the magnitude of differences between neighboring vectors in the DVF of the inverse transformation $T'$, using Hooke's law.
We denote the deformation vector at point $p_t$ as $T'_{p_t}$, to differentiate it from the deformed point $T'(p_t)$ (which would correspond to a point $p_s$).
Given all discrete voxel positions $p_t$ of the target image, and all neighbors $n(p_t)$ of each voxel position, this objective is defined as follows:
\[
f_{\text{\emph{def. magnitude}}} = \sum_{p_t \in P_t} \left[ \frac{1}{|n(p_t)|} \sum_{a \in n(p_t)} \lVert T'_{p_t} - T'_{a} \rVert^2 \right] 
\]

Using the newly calculated objective values, we then derive a new approximation set from each original set.
Due to the differing deformation magnitude objective formulations, certain non-dominated solutions in the original approximation set may become dominated, or vice versa.
A non-dominated solution can become dominated if there is another solution in the set that has a superior (or inferior) deformation magnitude objective value in the new objective formulation, and an equal or superior image similarity objective value.
Similarly, dominated solutions can become non-dominated if there is no other solution for which this condition holds.
In the results, solutions that become dominated in the new deformation magnitude objective formulation are visualized separately, in the interest of fairness.

To provide insight into registrations across each set, we highlight three registration solutions: The solution with the best deformation magnitude, the solution with the best image similarity, and an example trade-off solution.
The example trade-off solution is chosen automatically by normalizing all objective values to a unit square and choosing the solution of which the objective value vector forms an angle to the horizontal axis which is closest to 45 degrees.
This process selects a solution near the center of the front.

\input{figures/results/p1/fronts}
\input{figures/results/p2/fronts}
\input{figures/results/p3/fronts}

%% file: figures/illustrations/decomposition.tex
\newcommand{\decompsize}{4cm}
\newcommand{\decompfig}{4cm}
\newcommand{\decompgap}{0.3cm}

\begin{figure}
    \centering
    \subfloat[\label{fig:decomposition:spline-patch}\centering \textbf{B-spline:} The deformation of an image region (blue) is defined by 16 control points on the source image (black).]{\makebox[\decompsize][c]{\includegraphics[height=\decompfig,valign=c]{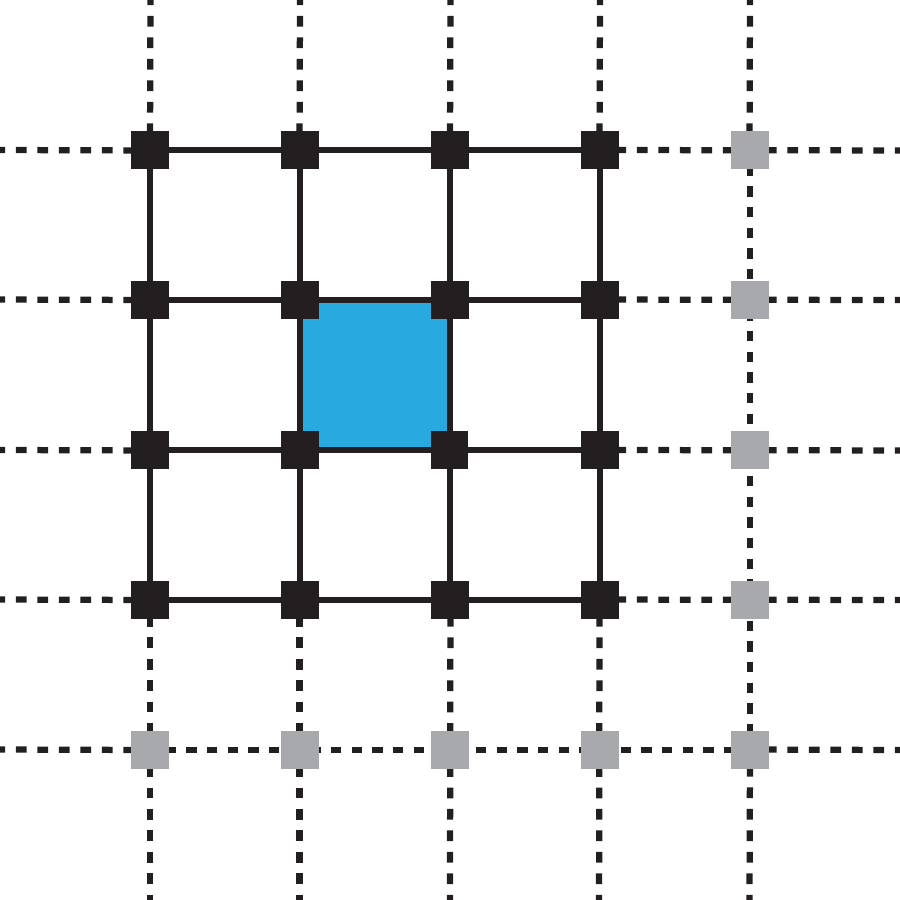}}}%
    \hspace{\decompgap}%
    \subfloat[\label{fig:decomposition:spline-dependencies}\centering \textbf{B-spline:} When one point (red) is moved, 25 points are considered to recompute 16 surrounding regions (blue).]{\makebox[\decompsize][c]{\includegraphics[height=\decompfig,valign=c]{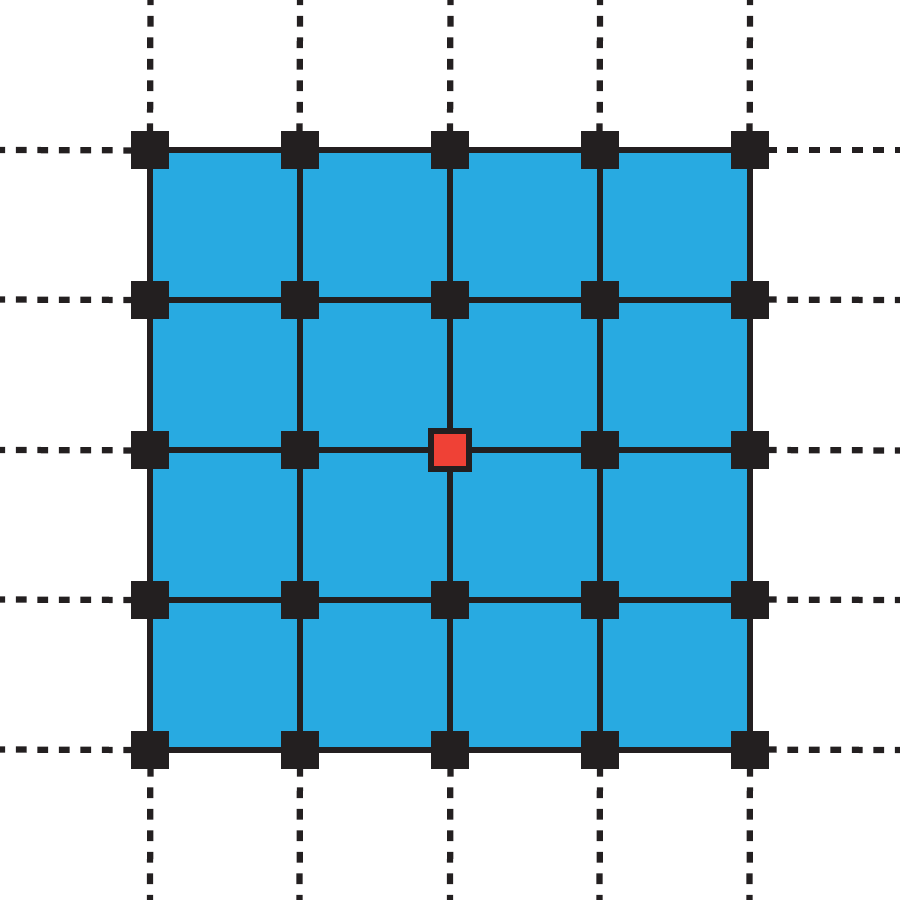}}}%
    \hspace{\decompgap}%
    \subfloat[\label{fig:decomposition:mesh-patch}\centering \textbf{Mesh:} The deformation of a triangular image region (blue) is defined by 6 mesh vertices, with 3 in each mesh (black).]{\makebox[\decompsize][c]{\includegraphics[height=\decompfig,valign=c]{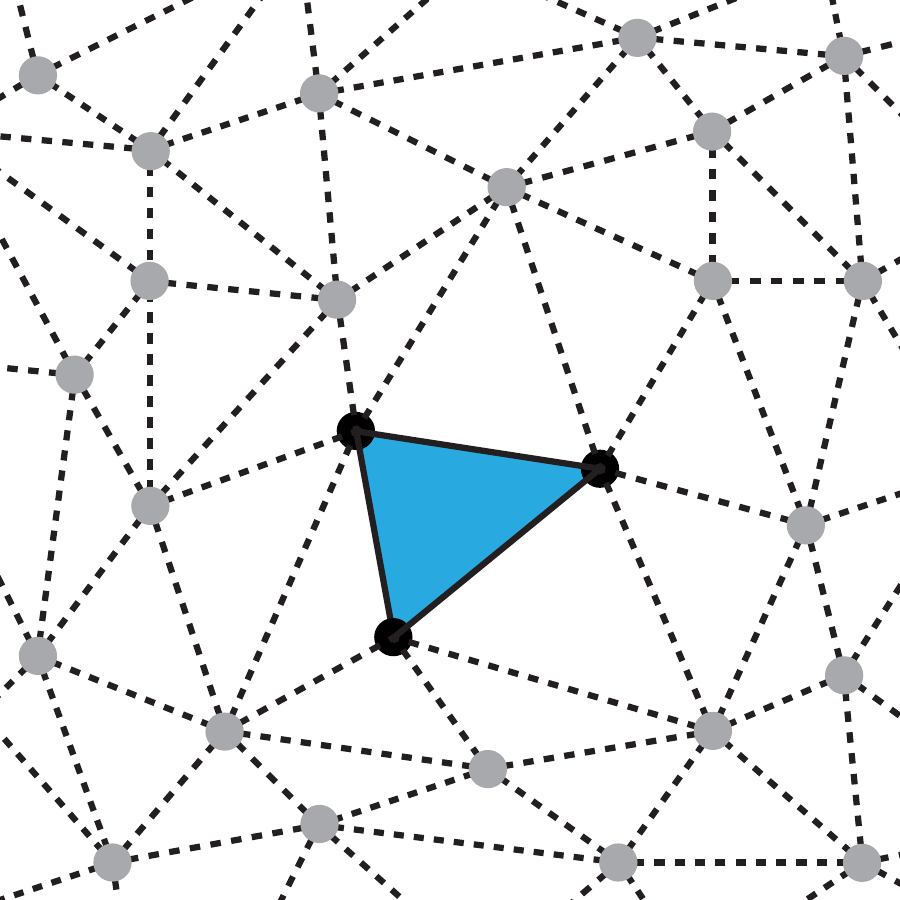}}}%
    \hspace{\decompgap}%
    \subfloat[\label{fig:decomposition:mesh-dependencies}\centering \textbf{Mesh:} When one vertex (red) is moved, vertices of surrounding triangles (blue) in both meshes are considered.]{\makebox[\decompsize][c]{\includegraphics[height=\decompfig,valign=c]{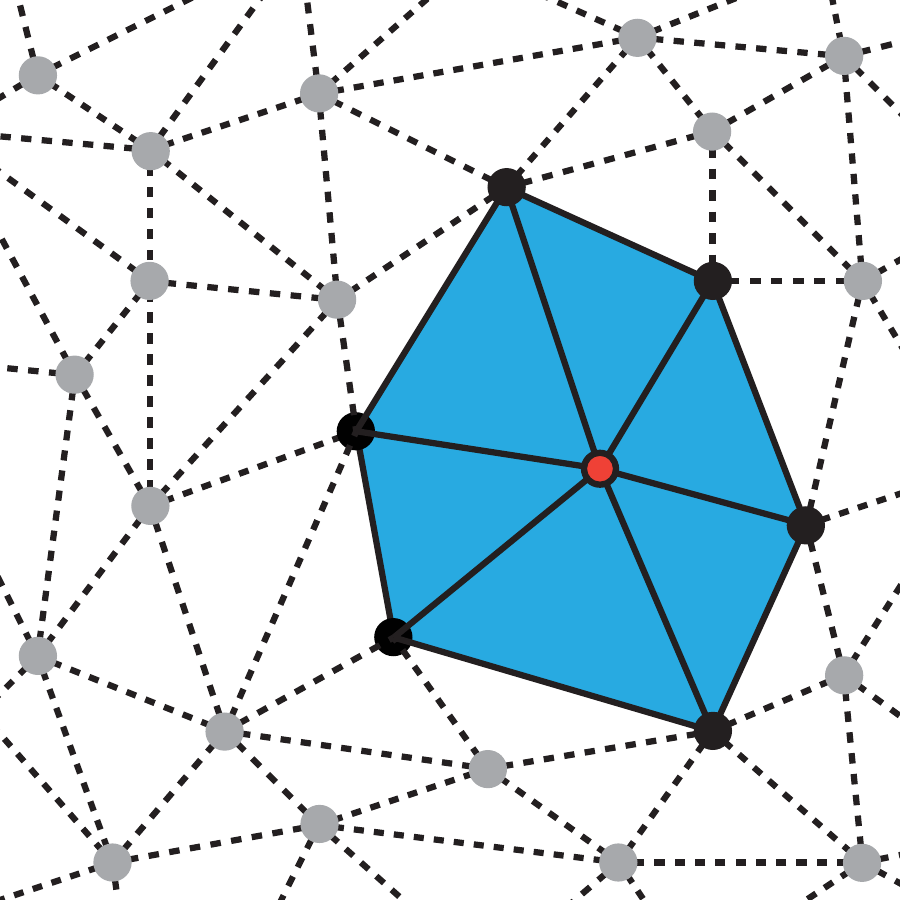}}}%
    \caption{2D illustration of how the B-spline transformation model and the dual-dynamic mesh transformation model both decompose the image domain and allow for partial, local objective value re-evaluations. Note that (a) and (b) visualize control point grids on the source image, while (c) and (d) visualize meshes on both the source and target image.}
    \label{fig:decomposition}
    \vspace{-0.1cm}
\end{figure}

%% file: figures/illustrations/optimization-methods.tex
\newcommand{\optsize}{5.5cm}
\newcommand{\optfig}{5.5cm}
\newcommand{\optgap}{0.3cm}

\begin{figure}
    \centering
    \subfloat[\label{fig:optimization-methods:b-spline-mo}\centering \textbf{B-spline multi-objective:} The B-spline transformation model is optimized using partial evaluations, making local changes to control point coefficients and evaluating their impact on their local region.]{\makebox[\optsize][c]{\includegraphics[width=\optfig,valign=c]{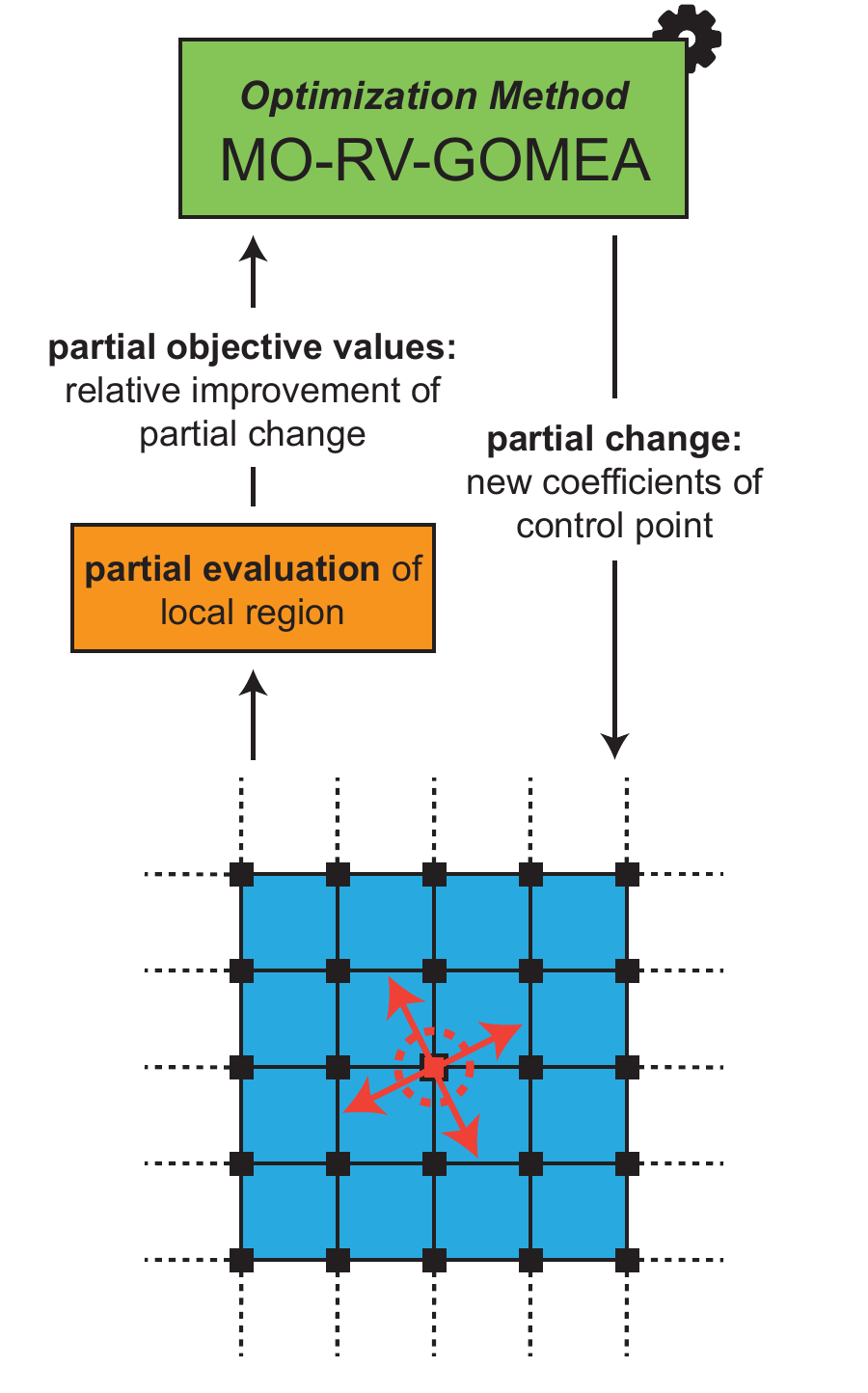}}}%
    \hspace{\optgap}%
    \subfloat[\label{fig:optimization-methods:mesh-mo}\centering \textbf{Mesh multi-objective:} The mesh transformation model is also optimized using partial evaluations, making changes to mesh node positions and evaluating their impact on their local region.]{\makebox[\optsize][c]{\includegraphics[width=\optfig,valign=c]{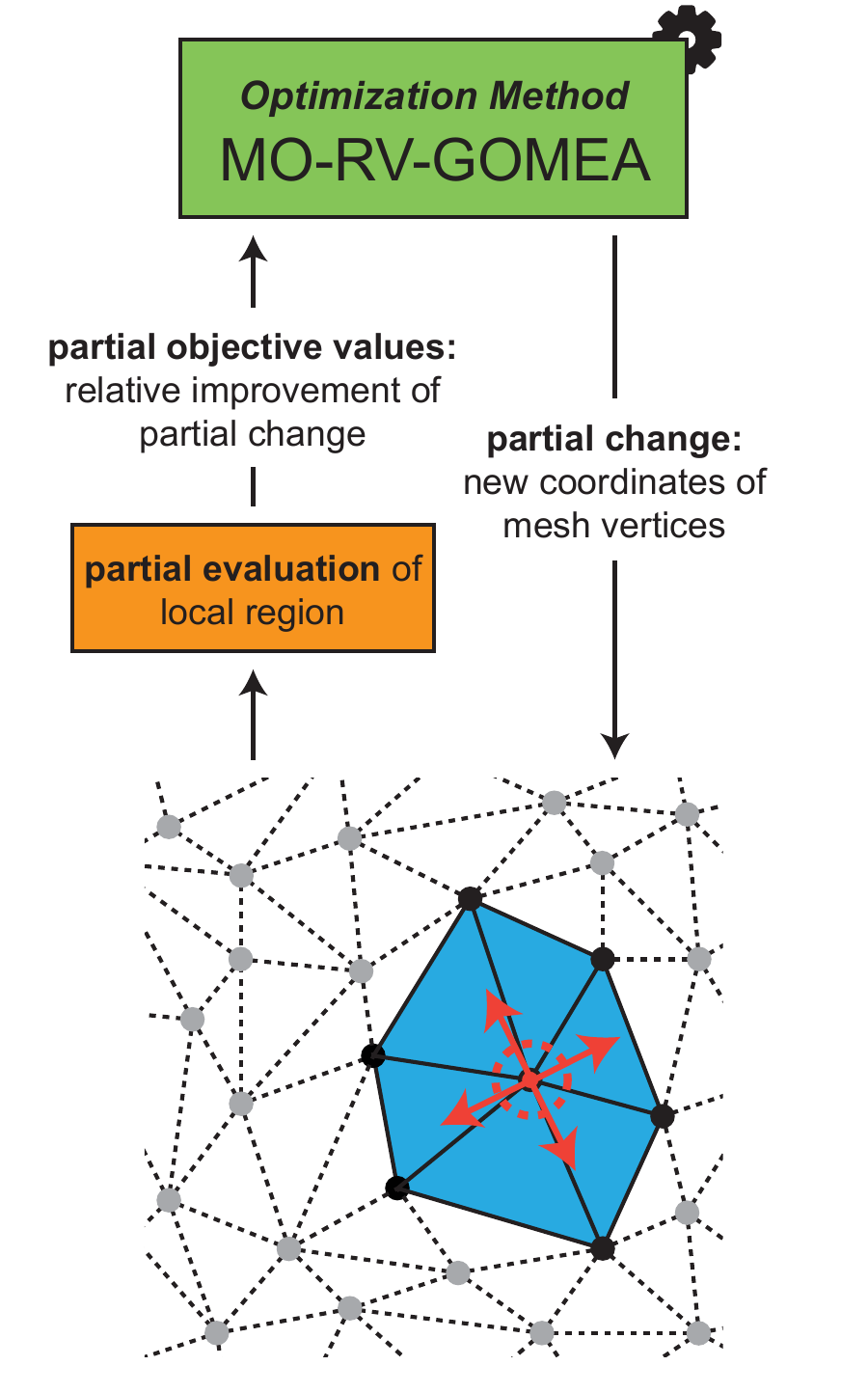}}}
    \hspace{\optgap}%
    \subfloat[\label{fig:optimization-methods:b-spline-baseline}\centering \textbf{B-spline baseline:} The comparison baseline optimizes the objective weights of a single-objective B-spline approach, instead of directly and multi-objectively optimizing the B-spline grid, itself.]{\makebox[\optsize][c]{\includegraphics[width=\optfig,valign=c]{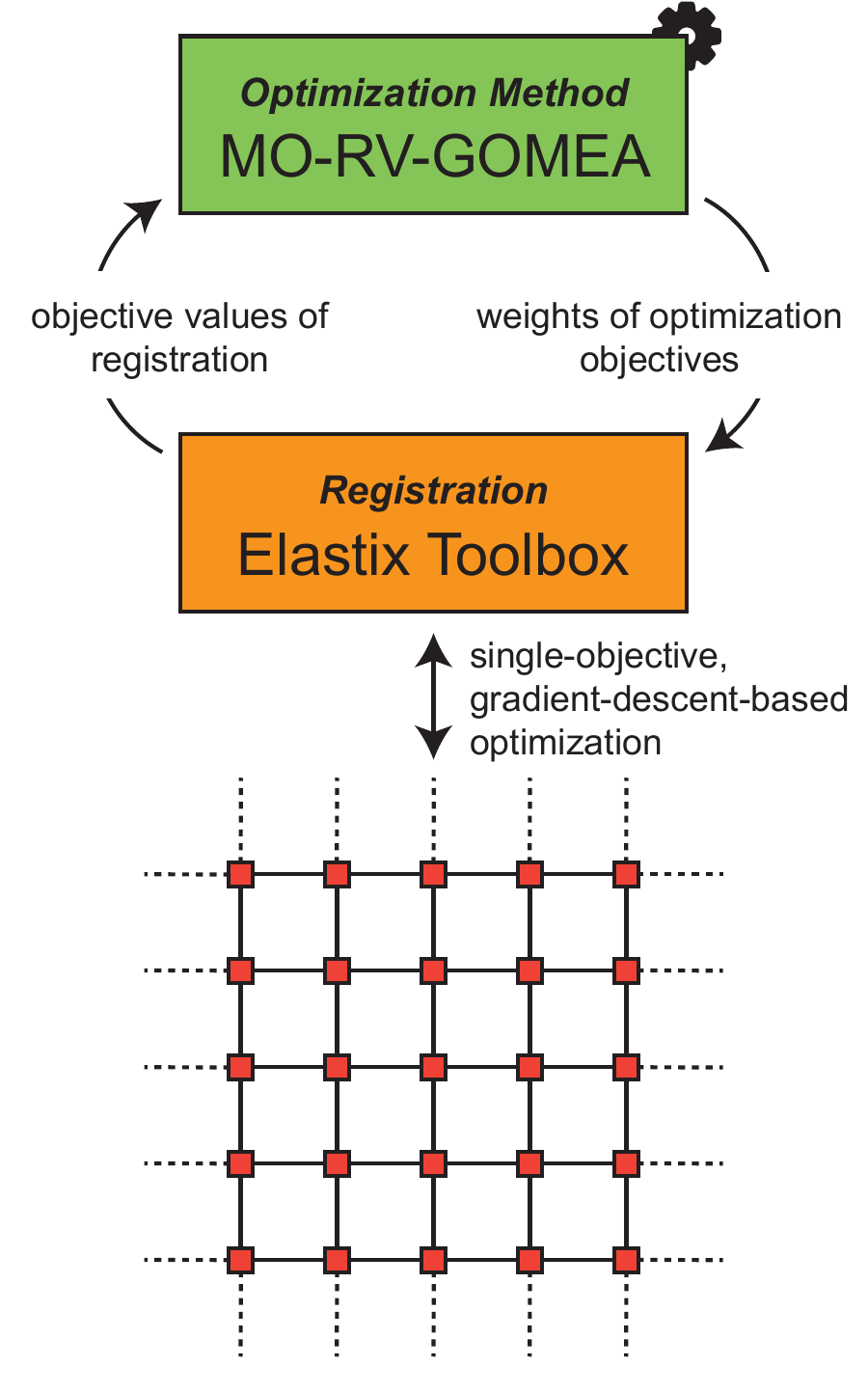}}}%
    \caption{2D illustration of the core optimization loops inside the three registration approaches which are experimentally compared in this work.}
    \label{fig:optimization-methods}
    \vspace{-0.1cm}
\end{figure}

%% file: figures/results/p1/problem.tex
\newcommand{\problemfig}{3.5cm}
\newcommand{\problemgap}{0.2cm}

\begin{figure}[t]
    \centering
    \subfloat[\label{fig:problem:bladder:source}\centering \textbf{Bladder:} Source]{\includegraphics[height=\problemfig,valign=c]{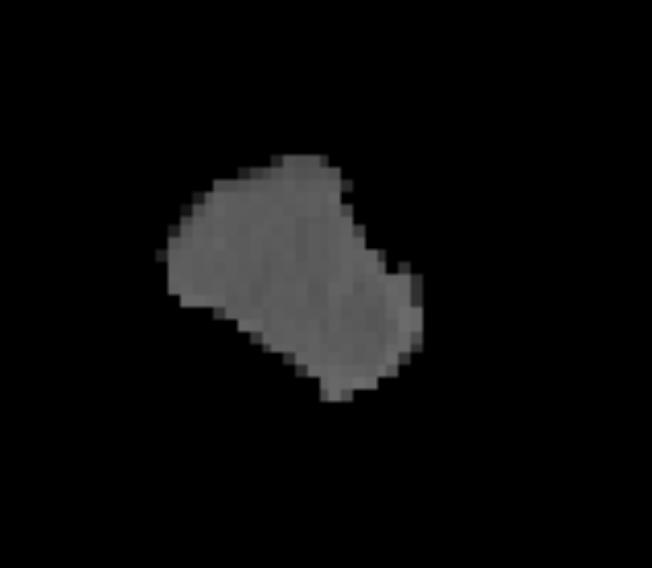}}%
    \hspace{\problemgap}%
    \subfloat[\label{fig:problem:bladder:target}\centering \textbf{Bladder:} Target]{\includegraphics[height=\problemfig,valign=c]{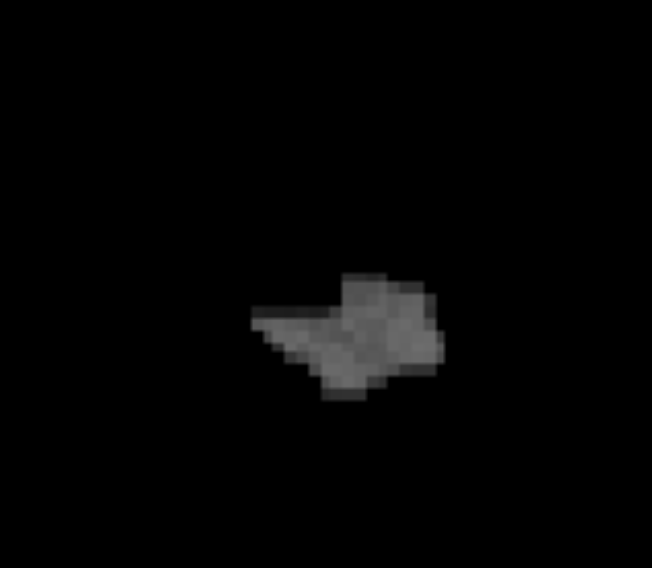}}%
    \hspace{\problemgap}%
    \subfloat[\label{fig:problem:patient:source}\centering \textbf{Patient:} Source]{\includegraphics[height=\problemfig,valign=c]{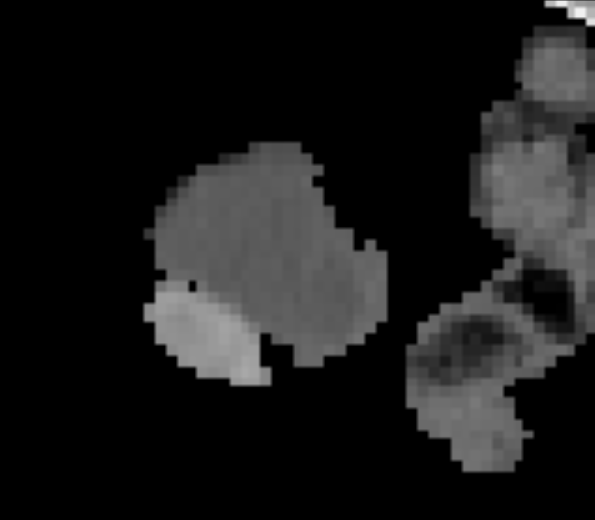}}%
    \hspace{\problemgap}%
    \subfloat[\label{fig:problem:patient:target}\centering \textbf{Patient:} Target]{\includegraphics[height=\problemfig,valign=c]{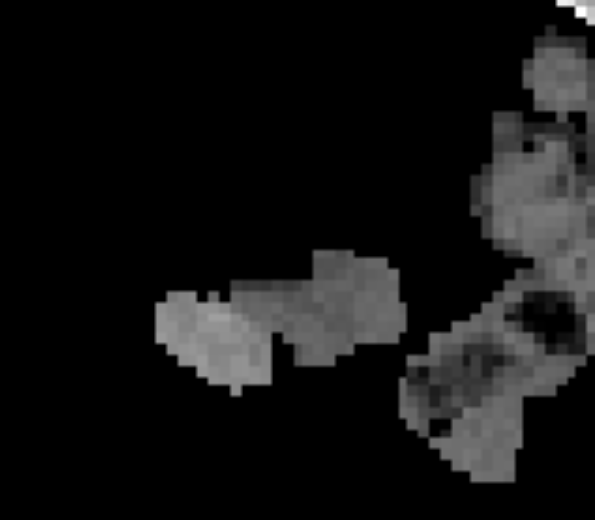}}
    \caption{Sagittal slices of the source and target images of the two registration problems of Patient 1.}
    \label{fig:problem}
    \vspace{-0.1cm}
\end{figure}

%% file: figures/results/p1/fronts.tex
\begin{figure}[t]
    \captionsetup[subfloat]{captionskip=-0.01cm}
    \centering
    \subfloat[\label{fig:fronts:p1:bladder}Registration problem with an isolated bladder.]{\includegraphics[width=\plotwidth\textwidth]{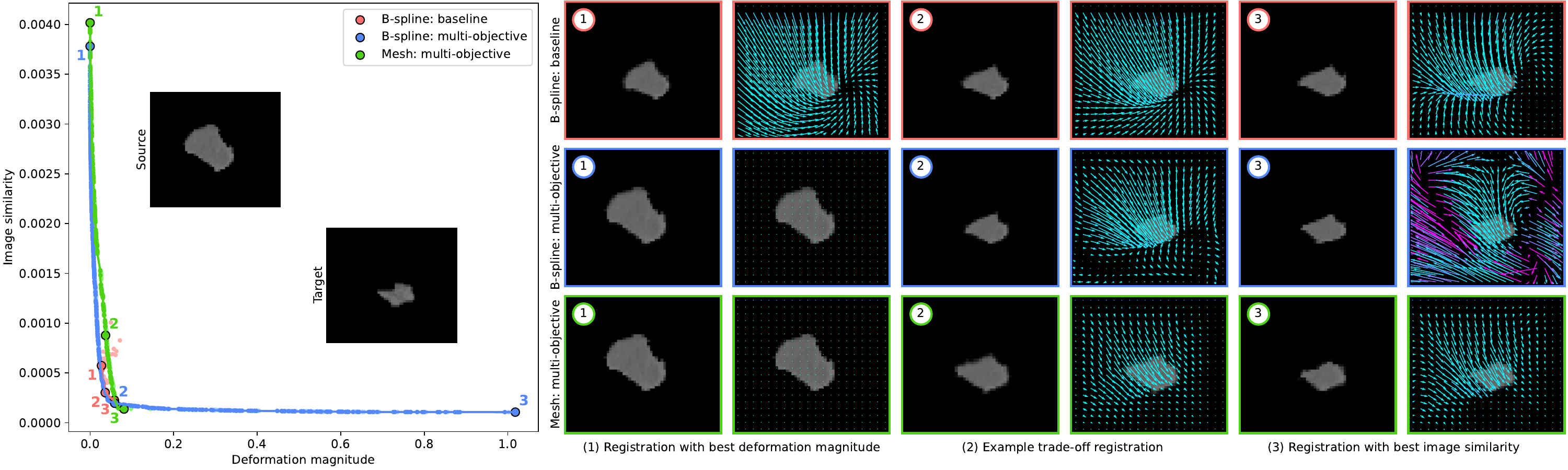}}\\
    \vspace{-0.2cm}
    \subfloat[\label{fig:fronts:p1:patient}Registration problem with multiple isolated organs.]{\includegraphics[width=\plotwidth\textwidth]{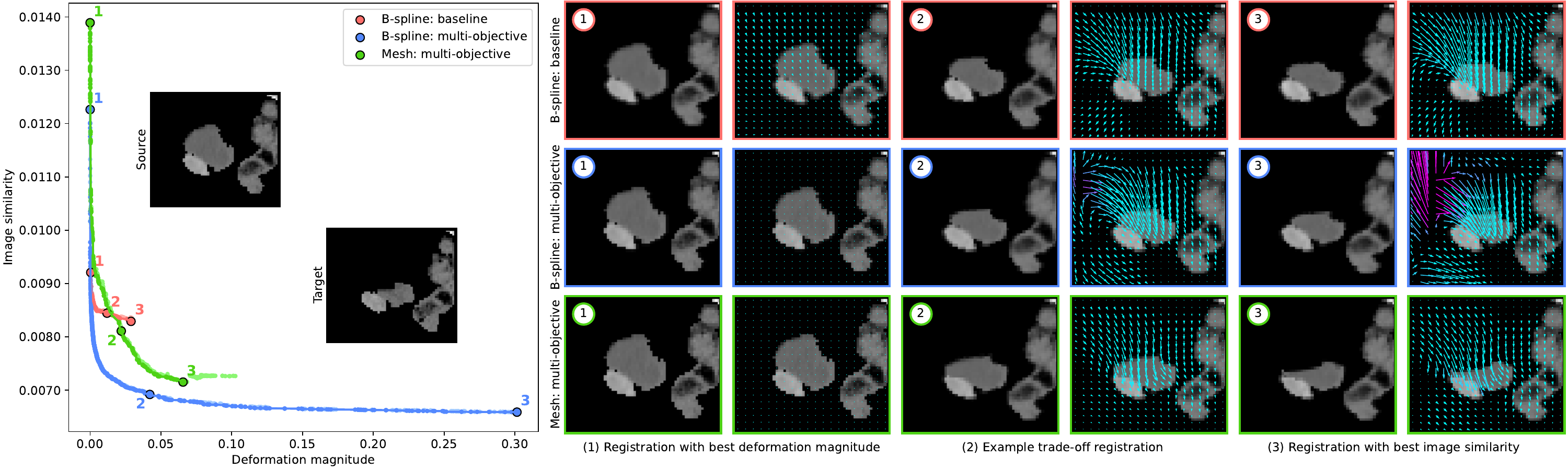}}
    
    \caption{Comparison of the registrations found by the three compared approaches for Patient 1. Each registration represents a different trade-off of the two optimization objectives. For each highlighted registration, the same sagittal slice of the transformed source image is shown and the DVF is rendered on top of the image, with arrow colors indicating local deformation magnitude objective values. Dominated solutions in each set of solutions are indicated with a lighter color shade.}
    \label{fig:fronts:p1}
    \vspace{-0.1cm}
\end{figure}

%% file: figures/results/p2/fronts.tex
\begin{figure}[t]
    \captionsetup[subfloat]{captionskip=-0.01cm}
    \centering
    \subfloat[\label{fig:fronts:p2:bladder}Registration problem with an isolated bladder.]{\includegraphics[width=\plotwidth\textwidth]{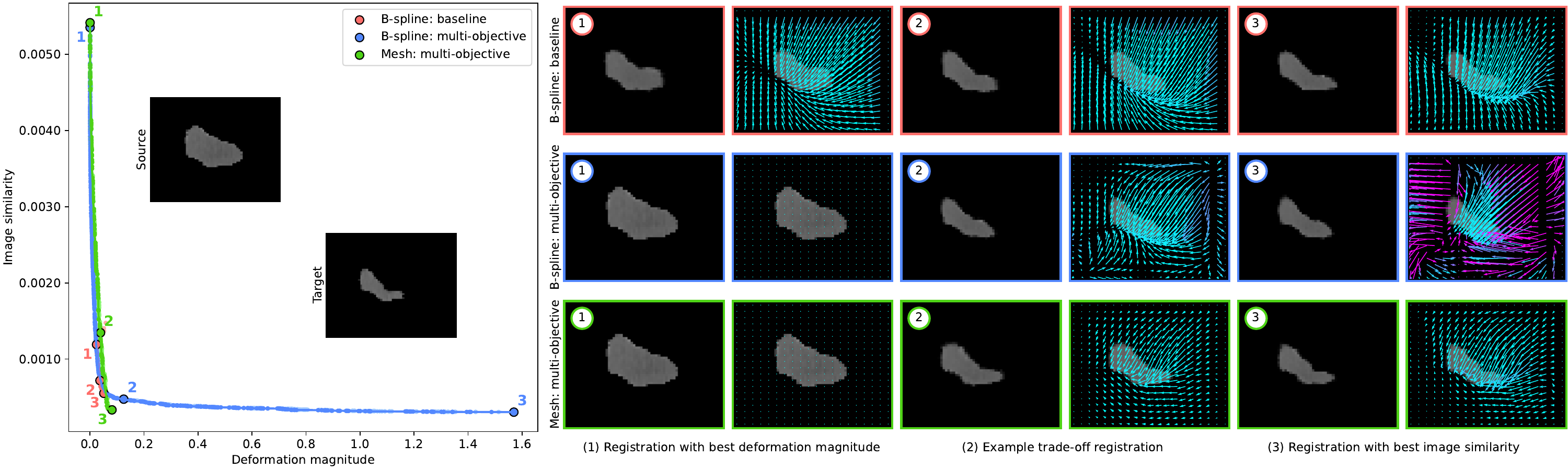}}\\
    \vspace{-0.2cm}
    \subfloat[\label{fig:fronts:p2:patient}Registration problem with multiple isolated organs.]{\includegraphics[width=\plotwidth\textwidth]{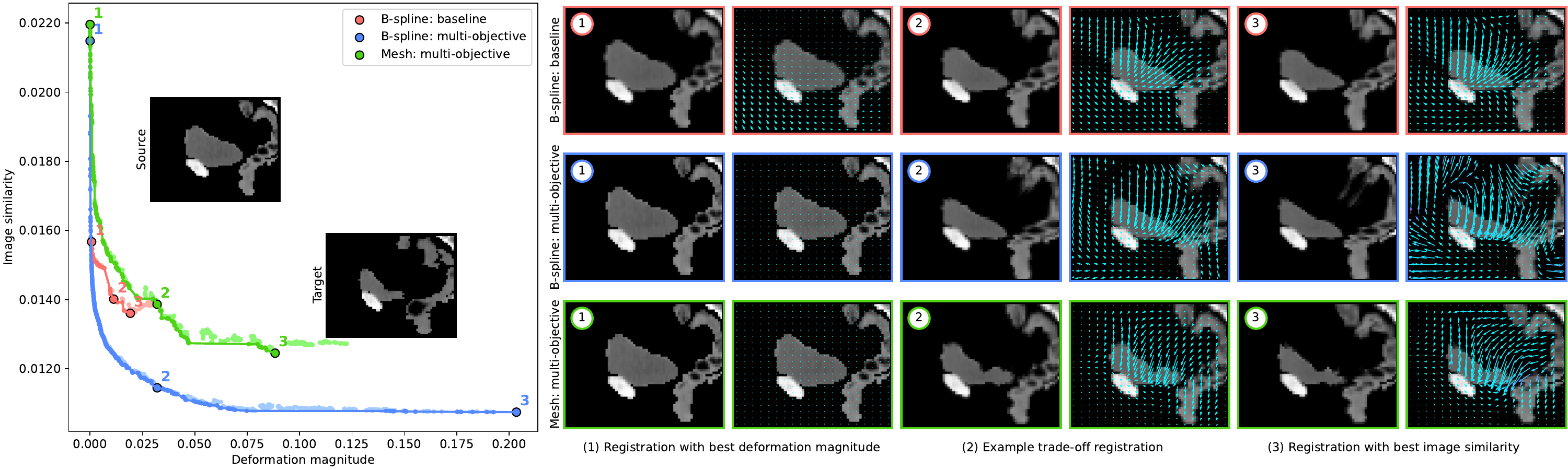}}
    
    \caption{Comparison of the registrations found by the three compared approaches for Patient 2. Further explanation is given in the caption of Figure~\ref{fig:fronts:p1}.}
    \label{fig:fronts:p2}
    \vspace{-0.1cm}
\end{figure}

%% file: figures/results/p3/fronts.tex
\begin{figure}[t]
    \captionsetup[subfloat]{captionskip=-0.01cm}
    \centering
    \subfloat[\label{fig:fronts:p3:bladder}Registration problem with an isolated bladder.]{\includegraphics[width=\plotwidth\textwidth]{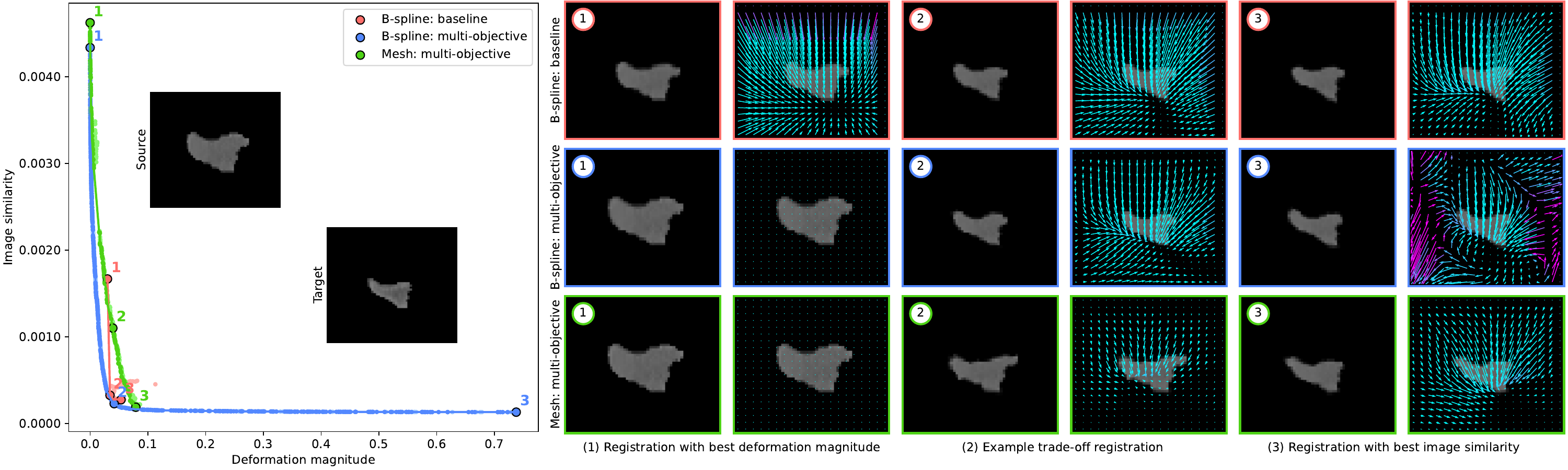}}\\
    \vspace{-0.2cm}
    \subfloat[\label{fig:fronts:p3:patient}Registration problem with multiple isolated organs.]{\includegraphics[width=\plotwidth\textwidth]{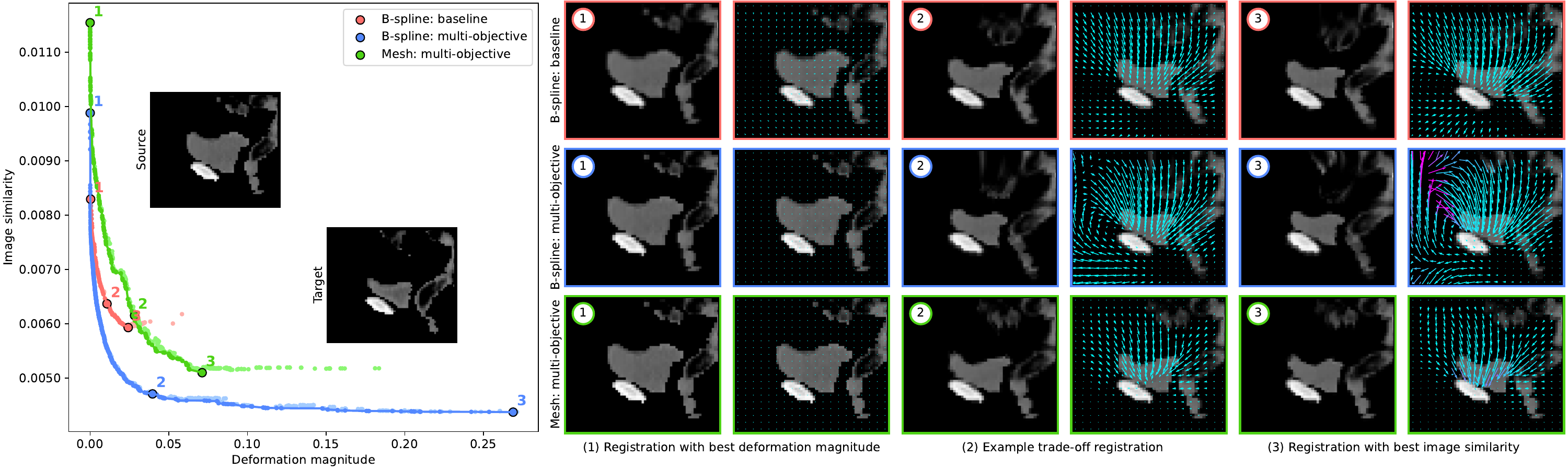}}
    
    \caption{Comparison of the registrations found by the three compared approaches for Patient 3. Further explanation is given in the caption of Figure~\ref{fig:fronts:p1}.}
    \label{fig:fronts:p3}
    \vspace{-0.1cm}
\end{figure}

%% file: sections/3-results.tex
\section{RESULTS}

For each of the three patients, corresponding with two registration problems each, we plot the objective values of the registrations found by all approaches in Figure~\ref{fig:fronts:p1}, \ref{fig:fronts:p2}, and~\ref{fig:fronts:p3}.
The three solutions selected from each approximation set are highlighted in the plot and visualized using a deformed source image and an overlay of a forward DVF\footnote{This forward DVF was computed by inverting the inverse DVF produced by each approach (originally defined in target image space).}.
First, we compare the registration approach introduced in this work, which directly optimizes a B-spline transformation model using MO-RV-GOMEA, to the baseline gradient-descent B-spline approach, which executes repeated single-objective runs and multi-objectively optimizes the objective weights.
We observe on all problems that the choice of optimization method has a strong impact on the set of achieved registration outcomes.
This impact is most clearly visible in the quality of deformations at the top of the deformed bladder shape.
Overall, the diversity and quality of the registrations of the fully evolutionary approach appear superior to the gradient-based baseline approach, presenting a wider range of higher quality solutions to the user. 
This can be seen, for example, in Figure~\ref{fig:fronts:p1:bladder}, where the horizontal and vertical spread of the baseline front (depicted in \mosoelastix) is much smaller than the approach in which MO-RV-GOMEA directly optimizes the model, itself (depicted in \moelastix).
We conclude that direct optimization of the model with MO-RV-GOMEA is a more powerful optimization approach in this setting.

When comparing the two transformation models directly, distinct differences can be observed in both registration problems of all three patients.
In the isolated bladder problem of Patient~1 (Figure~\ref{fig:fronts:p1:bladder}), the transformations of the mesh model (\morea) appear more localized, following the contours of the bladder more closely than the transformations found based on using the B-spline model (\moelastix).
Although the B-spline model provides a more diverse set of transformations, in terms of both objective values, its deformations display less locality: they require larger deformations in the entire image space to facilitate the deformation of the bladder.
This holds for highlighted registrations~2 and~3 of both transformation models, but is especially visible for registration~3, where the surrounding deformation of the B-spline model is especially strong.
These findings are corroborated by the isolated bladder problems of Patients~2 and~3 (Figures~\ref{fig:fronts:p2:bladder} and~\ref{fig:fronts:p3:bladder}, respectively), all of which show more localized transformations when registered with the mesh model (\morea).

In the problem with multiple organs of Patient~1 (Figure~\ref{fig:fronts:p1:patient}), we see that the results found by both approaches are more constrained by the image context.
The mesh-based approach (\morea) results in a more evenly distributed deformation across the bladder and again shows a more precise, localized deformation than the results obtained using the B-spline-based approach (\moelastix).
This can be seen in registration~3 of both transformation models.
Figure~\ref{fig:fronts:p2:patient} and~\ref{fig:fronts:p3:patient} of Patients~2 and~3 confirm this trend, with the mesh-based transformations approximating the target images more closely with more localized transformations.

%% file: sections/4-discussion-and-conclusions.tex
\section{DISCUSSION AND CONCLUSIONS}

For the first time, a B-spline transformation model has been directly compared to a mesh transformation model, using the same multi-objective optimization method.
This was made possible by introducing a novel registration approach that spatially decomposes the B-spline model, using the MO-RV-GOMEA optimization method.
Our experimental comparison between this B-spline approach and the mesh-based MOREA approach shows profound differences in diversity, quality, and locality of registrations produced by the different transformation models, even if optimized with the same optimization method.
These findings emphasize the need for a deliberate choice of transformation model for each DIR application, as this choice can strongly impact the realism of transformations.
Moreover, our findings stress the need for multi-objective registration, as the shapes of individual approximation fronts vary strongly between problem instances.
This makes it difficult to choose a general set of objective weights for single-objective approaches, upfront.
The outcomes of a single-objective approach may therefore be less stable across instances, as it is difficult to consistently target a certain part of the front without knowing the shape of each approximation front.

There are limitations to the validity of this comparison.
For example, we chose to run all approaches in single-resolution mode, as some approaches were not sufficiently hardware-accelerated to support multi-resolution registration. 
It is unclear what the impact of a multi-resolution mode would have on the outcome of this comparison.
Another limitation is the resolution of the images used, which was chosen to be low enough to be computationally tractable for all approaches.
As important image features remained visible at this resolution and all approaches were given the same image resolution, this should however not obstruct the validity of this comparison.
A third limitation arises from the use of a different deformation magnitude metric for the post-hoc comparison between models. 
This different formulation may correlate more strongly with the B-spline formulation than with the mesh formulation, since it considers transformation gradients on a voxel-by-voxel basis.
This puts the B-spline model at an advantage in the post-hoc comparison.
Finally, the choice of objectives may be of influence.
We chose a subset of objectives and metrics that both models supported, but this excludes the use of more advanced objectives such as contour-based guidance (which MOREA supports) or metrics such as mutual information image similarity (which Elastix supports).

Future research should aim to address the abovementioned limitations.
To further validate the findings of this study, blind expert assessments of the registrations found by different models could be conducted.
Moreover, this empirical comparison could be extended to other transformation models, such as radial basis function approaches~\cite{Rohde2003} or kernel-based stationary velocity fields~\cite{Pai2015}.
A similar comparison could be made on registration problems with small deformations, where different considerations may be relevant.